%% file: acl2024.tex
\pdfoutput=1

\documentclass[11pt]{article}

\usepackage[]{ACL2024}

\usepackage{times}
\usepackage{latexsym}

\usepackage[T1]{fontenc}

\usepackage[utf8]{inputenc}

\usepackage{microtype}

\usepackage{inconsolata}
\usepackage{amsmath}
\usepackage{graphicx}
\usepackage{multirow}
\usepackage{latexsym}
\usepackage{amssymb}
\usepackage{makecell}
\usepackage{booktabs}
\usepackage{bbm}
\usepackage{wrapfig}
\usepackage{lipsum} 
\usepackage{xcolor}
\usepackage{caption}

\def\ours{RaFe}

\title{RaFe: Ranking Feedback Improves Query Rewriting for RAG}

\author{Shengyu Mao$^\spadesuit$, Yong Jiang$^\heartsuit$\thanks{$\quad$ Corresponding Author.}, Boli Chen$^\heartsuit$, Xiao Li$^\diamondsuit$, Peng Wang$^\spadesuit$, 
Xinyu Wang$^\heartsuit$,\\ \textbf{Pengjun Xie$^\heartsuit$, Fei Huang$^\heartsuit$, Huajun Chen$^\spadesuit$, Ningyu Zhang$^{\spadesuit*}$}\\
$^\spadesuit$Zhejiang University~
$^\heartsuit$Alibaba Group,~
$^\diamondsuit$Nanjing University
\\
\texttt{\{shengyu,zhangningyu\}@zju.edu.cn}~,
\texttt{yongjiang.jy@alibaba-inc.com}
}

\begin{document}
\maketitle
\begin{abstract}

As Large Language Models (LLMs) and Retrieval Augmentation Generation (RAG) techniques have evolved, query rewriting has been widely incorporated into the RAG system for downstream tasks like open-domain QA. Many works have attempted to utilize small models with reinforcement learning rather than costly LLMs to improve query rewriting. However, current methods require annotations (e.g., labeled relevant documents or downstream answers) or predesigned rewards for feedback, which  lack generalization, and fail to utilize signals tailored for query rewriting. In this paper, we propose \ours, a framework for training query rewriting models free of annotations. By leveraging a publicly available reranker,  \ours~provides feedback aligned well with the rewriting objectives. Experimental results demonstrate that \ours~can obtain better performance than baselines.

\end{abstract}

\section{Introduction}

\begin{figure}
    \centering
    \includegraphics[width=1\linewidth]{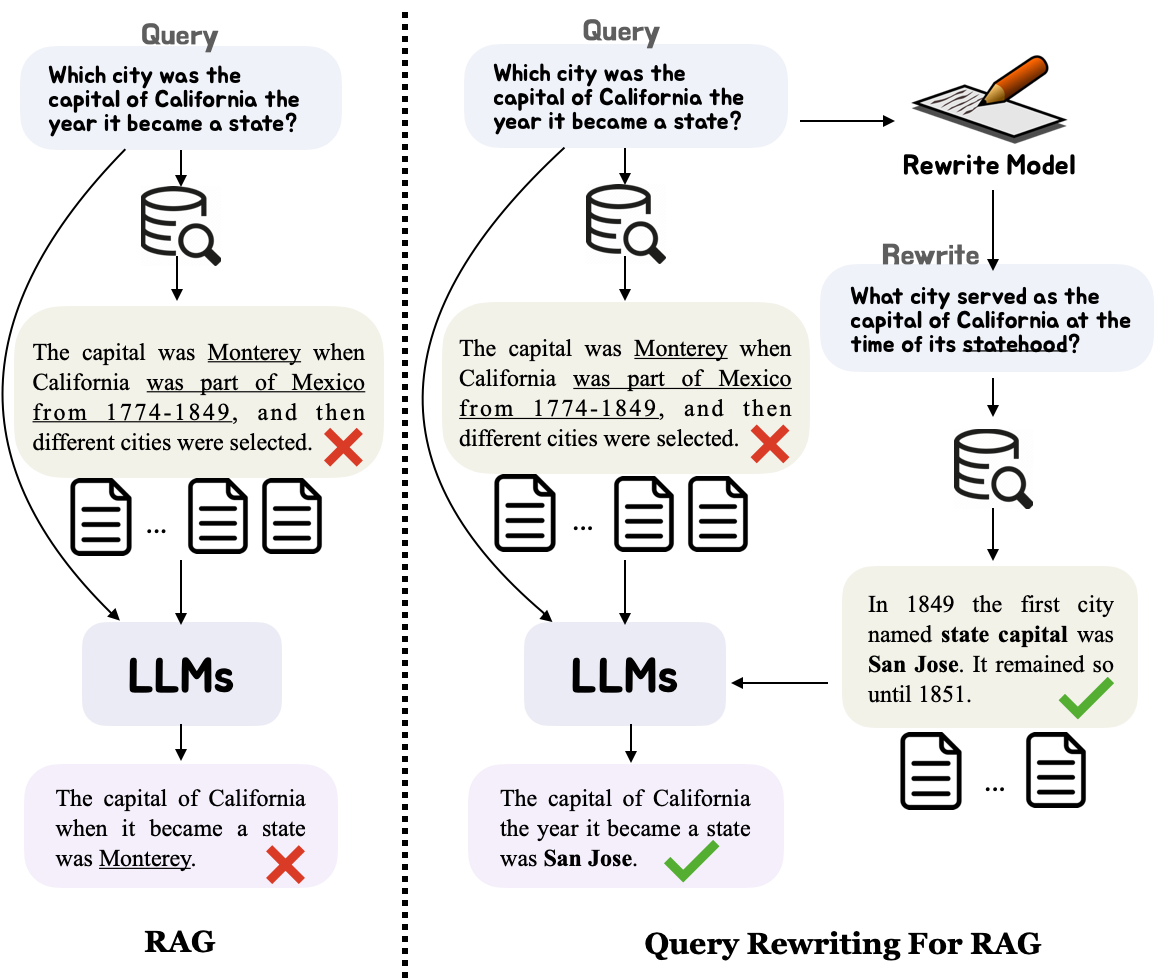}
    \caption{Illustration of query rewriting for RAG.
    The left part indicates the normal RAG pipeline, while the right part presents the query rewriting to expand more relevant documents for RAG.}
    \label{fig:fig_1}
\end{figure}

Large Language Models~(LLMs) have demonstrated strong capacities to solve a variety of tasks \cite{DBLP:journals/corr/abs-2303-18223}.
However, they still encounter the challenges of hallucinations~\citep{DBLP:journals/csur/JiLFYSXIBMF23, DBLP:journals/corr/abs-2309-01219, DBLP:journals/corr/abs-2311-05232} or outdated knowledge~\citep{DBLP:conf/emnlp/YaoWT0LDC023,DBLP:journals/corr/abs-2401-01286}. 
Recently, Retrieval Augmentation Generation (RAG)~\citep{RAG_Survey} has become an important technology to enhance LLMs' abilities, by incorporating external knowledge.
For instance, in open-domain QA, LLMs can firstly retrieve related documents and then generate answers.
Nonetheless, directly retrieving by original query does not always achieve correct and relevant documents.
Therefore, query rewriting~\citep{efthimiadis1996query, DBLP:journals/csur/CarpinetoR12} has been widely employed to reformulate the query to expand the retrieved documents for a better response as illustrated in Figure~\ref{fig:fig_1}.

Many efforts have been proposed to leverage the powerful LLMs to directly generate rewrites~\citep{lamer, query2doc}.
While in practical applications, it is more prevalent to implement specific small query rewriting models to avoid the costly use of LLMs~\citep{QueryRewrite_emnlp23}. 
To improve the performance of query rewriting, reinforcement learning (RL) with feedback ~\citep{ConQRR, reinforced_question_rewrite}  can be utilized as a typical solution.
For instance, \citet{query_reformulation} generates feedback by considering the recall of labeled documents. 
Meanwhile, \citet{QueryRewrite_emnlp23} leverages evaluation results from question answering (QA) post-rewriting to generate signals. 
Additionally, \citet{taobao_rewrite} employs domain-specific annotated rewriting scores for feedback training.

Note that these feedback-driven query rewriting methods rely on either annotated labels such as relevant documents or answers, or pre-designed rewards tailored to specific domains. 
However, they often lack the utilization of effective and general signals for query rewriting.
Meanwhile, considerable efforts have been made to harness diverse feedback mechanisms across various domains~\citep{maf, tuna}. 
Notably,  ~\citet{RLTF} effectively integrates unit testing feedback into code generation, yielding significant efficacy.
Drawing from these, in this paper we attempt to 
(i) reduce the \textbf{cost of annotations for feedback}; 
and (ii) identify \textbf{a signal that better aligns with the objectives of the query rewriting} task.


To address these issues, we introduce \textbf{\ours} (\textbf{Ra}nking \textbf{Fe}edback improves Query Rewriting), a novel framework that leverages feedback from the reranker to train query rewriting models.
This approach is inspired by the reranker module in traditional information retrieval (IR) systems, which score and sort retrieved documents based on the query.
Intuitively, query rewriting aims to retrieve documents relevant to the original query, which aligns perfectly with the goal of the reranker. Specifically, the reranker is capable of scoring documents without requiring additional labels. 
Thus, we incorporate a reranker to provide feedback for the query rewriting model.

\ours~comprises a two-stage process. 
We first train an initial query rewriting model by standard supervised fine-tuning.
Subsequently, we utilize the ranking scores from the reranker to conduct feedback training on the query rewriting model.
\ours~supports both offline and online RL feedback training.
Empirically, we demonstrate that utilizing a general, publicly available reranker, \ours~can drive the training of the query rewriting model, indicating the effectiveness and potential generalizability of the proposed approach. 
The main contributions of our paper can be summarized as follows:
\begin{itemize}
    \item We propose~\ours, a novel query rewriting framework that utilizes feedback from the reranker, an especially fitting signal for the objective of retrieving more relevant documents.
    \item \ours~does not necessitate annotated labels or particularly designed scores, ensuring the generalizability of the training framework.
    \item We validate the effectiveness of our proposed approach on cross-lingual datasets across wide settings with a general and public reranker, we further conduct a comprehensive investigation of what makes a better query rewriting and how ranking feedback works.
\end{itemize}

\begin{figure*}
    \centering
    \includegraphics[width=1\linewidth]{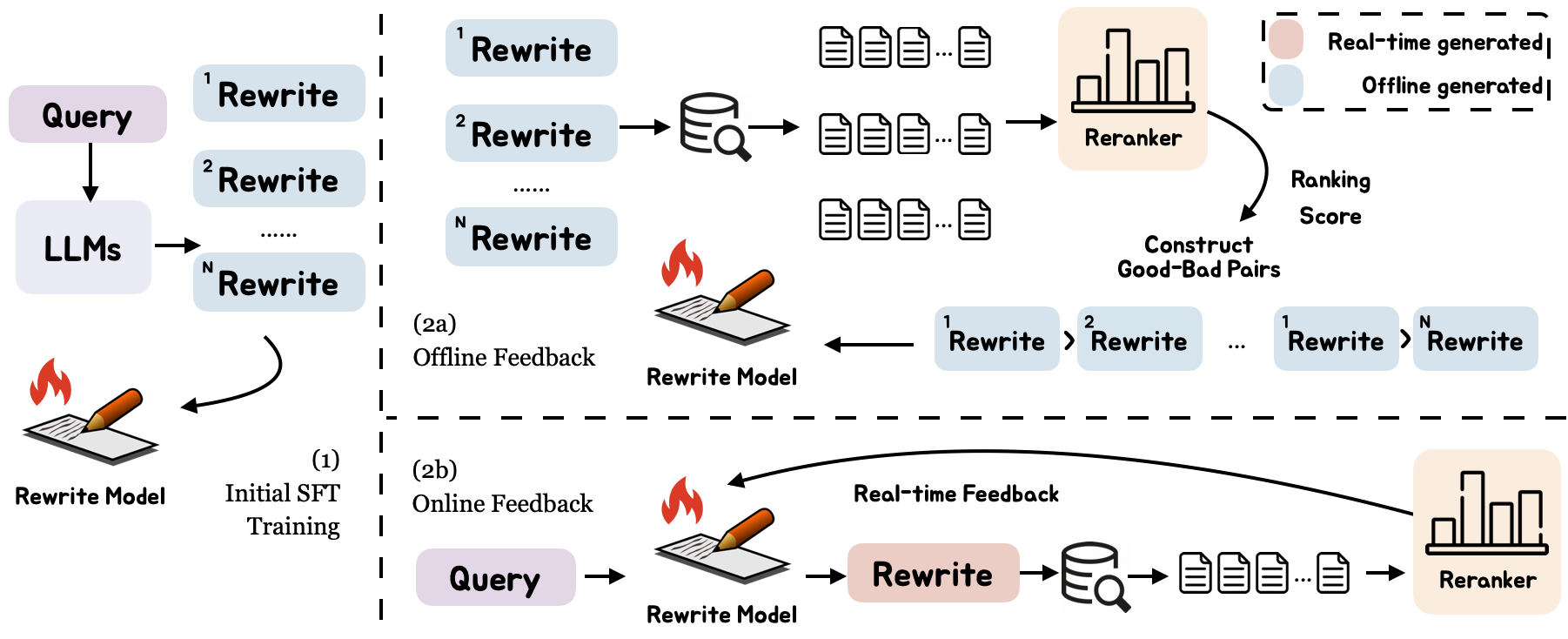}
    \caption{The overview of \textbf{RaFe}. The entire procedure consists of two stages: 
    the initial SFT, and subsequent feedback training. 
    RaFe obtains ranking feedback aligned with the goal of query rewriting without annotated data
    and enables leveraging the feedback in two ways. 
    \textbf{Offline training}: Constructing good-bad pairs from offline-generated data. 
    \textbf{Online training}: Scoring queries generated in real-time and complete feedback training.
    }
    \label{fig:main}
\end{figure*}

\section{Method}
\input{sec/method}

\section{Experimental Setup} \label{sec:exp}
\input{sec/experiment}

\section{Results}
\input{sec/result}

\section{Analysis}
\input{sec/analysis}

\section{Related Work}
\input{sec/related_work}

\section{Conclusion and Future Work}
This paper proposes a novel feedback training framework named \textbf{\ours} for query rewriting, 
based on the effectiveness of the reranker in enhancing document ranking during the information retrieval process.
By leveraging the feedback signals from reranker, \ours~is capable of effectively and generally conducting feedback training for rewrite models, yielding great improvements.
Experimental results indicate that our method achieves exemplary performance across cross-linguistic datasets. 
In the future, we plan to conduct the joint training of reranker and rewrite models, which may yield substantial benefits for RAG.
\section*{Limitations}
Although our experiments employ a general reranker as the source of feedback signals, there are still some limitations. 
(1) The Lack of Cross-Domain Validation. As constrained by the lack of domain-specific data, we lack the validation of separately trained rerankers on datasets pertinent to a specific domain.
(2) Reliance on the Effectiveness of Rewriting as a Bottleneck. Although we can achieve some improvements by using publicly available rerankers, this enhancement may be limited by the capability of the reranker.



\bibliography{anthology,custom}
\bibliographystyle{acl_natbib}

\appendix
\section{Appendix}
\input{sec/appendix}

\end{document}

%% file: sec/method.tex
\subsection{Task Formulation}

Within the process of Retrieval Augmented Generation~(RAG), when inputting an original query $q$, a set of relevant documents $D=[d_0,d_1,...,d_k]$ will be retrieved through a search engine, and the retrieved documents are utilized to enable the model to better accomplish the corresponding task (in this paper, we discuss the task of Open-domain Question Answering).
Query rewriting is to reformulate the original query $q$ into another form to better retrieve relevant passages. 
We aim to obtain a better rewrite model $\mathcal{M_{\theta}}$ that can rewrite $q$ as:
\begin{equation}
    q' = \mathcal{M_{\theta}}(q),
\end{equation}
here $q'$ is the rewritten query which is used to retrieve documents $D'$ for completing subsequent task.
Figure~\ref{fig:main} shows the overview of our proposed framework, \ours~ for query rewriting training.

\subsection{Initial Supervised Fine-Tuning}
Before leveraging the ranking feedback, we first initialize the rewrite model with a cold start supervised fine-tuning to gain the rewrite ability.
Specifically, we prompt the LLMs to produce the rewrite data, 
The details of the datasets we used to produce the training rewrite can be found in Sec~\ref{sec:dataset}.
The rewrites generated from LLMs are denoted as $T_{\text{all}}=\{(q, q')|q' \in Q'\}$, 
here 
$Q'$ is the rewrite set of original query $q$.
We split the training instances into two parts $T_{\text{all}} = [T_{\text{sft}}: T_{\text{f}}]$, here $T_{\text{sft}}$ and $T_{\text{f}}$ indicates the instances we use for SFT and feedback training, respectively.
We train the rewrite model $\mathcal{M}_{\theta}$ with standard SFT loss as follows:

\begin{equation}
    \mathcal{L}_{\text{sft}} = -\sum\nolimits_{{q'}\in {Q'}}\sum\nolimits_{t}  \log\mathcal{M_\theta}({q_{t}'}|{q_{<t}'}, q)
\end{equation}

Note that for each query, we mix all corresponding rewrites together in the dataset for training, to enhance the diversity of generation by our trained model, since in real-world applications, different rewrites are required for a single search query to address different aspects or interpretations.

\subsection{Feedback Training}\label{sec:feedback training}


 The evaluation of query rewriting is notoriously difficult due to the absence of direct quality assessment methods~\citep{llm_for_ir_survey}, so previous feedback for QR typically rely on the annotated passages~\citep{query_reformulation, ConQRR}.
While throughout the traditional IR pipeline, documents expanded by query rewriting are typically subjected to a reranking process. 
Intuitively, the reranker can serve as a natural feedback for query rewriting. 
Given a reranker model $\mathcal{M}_{r}$, the process of scoring a document $d$ with query $q$ can be formulate as $\mathcal{M}_{r}(q, d)$.
The ranking score of a rewrite $q'$ can be denoted as follow:

\begin{equation}
    S(q, q') = \frac{1}{|D'|}\sum\nolimits_{d'\in D'}
                             \mathcal{M}_{r}(q, d').
\end{equation}

In this way, we can provide reliable feedback for training rewriting models. 
As illustrated in Figure~\ref{fig:main}, our proposed method can be applied for both offline and online feedback training.

\input{tab/main_table_raw}

\paragraph{Offline Feedback}
For offline feedback, we leverage the ranking score of each documents retrieved by rewritten query to construct the preference data. Specifically, we set a threshold to distinguish the good and bad rewrites formulated as $\mu$, which is computed as the average ranking score for all training instances as follows:

\begin{equation}
    \mu = \frac{1}{|T_{\text{f}}|}\sum\nolimits_{(q, q')\in T_{\text{f}}}
                  S(q, q').
\end{equation}

Then for every rewrite $q'$ with a score exceeding the threshold $\mu$,  we regard it as a good rewrite for the original query $q$; otherwise, it is deemed a bad rewrite. In this way, we obtain all the preference pairs for open domain QA in the form $(q, q'_{g}, q'_{b})$.

For the offline feedback training, we use DPO~\citep{dpo} and KTO~\citep{kto}. 
DPO directly leverage the preference pairs to optimize the model, 
while KTO is a method that can optimize the model from feedback, only needs the signal of whether a rewrite $q^{'}$ is good or not, rather than needing pairs, formulated as $(q,q^{'};\rho), \rho\in[\text{good, bad}]$. 
The specific formulation of $\mathcal{L}_{kto}$ is in Eq~\ref{eq:kto_loss}, 
and the detailed explanation of the KTO is demonstrated in Appendix~\ref{appendix:training}.

\paragraph{Online Feedback}

The ranking score can also serve as an online feedback signal. 
We utilize the Proximal Policy Optimization (PPO)~\cite{ppo} algorithm to implement online feedback training.
The training process includes rewriting, retrieving, scoring and ultimately providing feedback, as illustrated in Figure~\ref{fig:main}(2b).
The details of the PPO loss and implementation are provided in Appendix~\ref{appendix:training}.

%% file: tab/main_table_raw.tex
\begin{table*}[ht]
\centering
\resizebox{2\columnwidth}{!}{
\begin{tabular}{l|cccccccc|cccc}

\toprule
\multirow{3}{*}[-0.5em]{\textbf{Method}} &  \multicolumn{8}{c|}{\textbf{EN}} & \multicolumn{4}{c}{\textbf{ZH}} \\
\cmidrule{2-13}
& \multicolumn{2}{c}{\textbf{FreshQA}} & \multicolumn{2}{c}{\textbf{NQ}} & \multicolumn{2}{c}{\textbf{TriviaQA}} & \multicolumn{2}{c|}{\textbf{HotpotQA}} & \multicolumn{2}{c}{\textbf{FreshQA}} & \multicolumn{2}{c}{\textbf{WebQA}} \\
& QA & Prec@5 & QA & Prec@5 & QA & Prec@5 & QA & Prec@5 & QA & Prec@5 & QA & Prec@5 \\

\midrule
w/o retrieval
& 41.70 & -     
& 43.74 & - 
& 74.99 & - 
& 34.80 & - 
& 40.98 & -
& 73.95 & -
\\

OQR               
& 61.87 & 27.48
& 51.36 & 32.35 
& 79.63 & 50.32 
& 42.75 & 17.73 
& 43.70 & 16.24 
& 81.29 & 77.25 
\\
\midrule
\multicolumn{13}{c}{\textsc{Substitute}-Raw} \\
\midrule

LLM-Rewrite             
& 57.38 & 25.23
& 48.62 & 29.83 
& 78.43 & 48.10 
& 40.92 & 15.32 
& 40.65	& 15.42
& 80.56	& 74.26
\\

Query2Doc              
& 56.52 & 26.08
& 46.12 & 27.65 
& 77.22 & 50.58 
& 38.85 & 16.26  
& 42.90	& 15.20
& \textbf{81.35}	& \underline{77.63}
\\

SFT$_{(T_{\text{sft}})}$                   
& 60.53 & 25.72
& 49.86 & 30.08 
& 78.34 & 47.77 
& 42.04 & 16.46 
& 42.44	& 15.56
& 77.76	& 72.65
\\

SFT$_{(T_{\text{all}})}$                 
& 60.55 & 24.88
& 50.39 & 30.40 
& 78.63 & 47.92 
& 42.66 & 16.89 
& 42.33	& 15.21
& 77.80	& 74.61
\\

RaFe$_{(PPO)}$                  
& \underline{62.21} & 27.72
& 50.83 & 31.52 
& 78.56 & 49.18 
& 43.82 & 17.64 
& 43.28	& 16.31
& \underline{81.28}	& \textbf{77.90}
\\

RaFe$_{(DPO)}$                 
& \textbf{62.67} & \underline{27.92}
& 51.14 & 32.25 
& \textbf{79.84} & \underline{50.67} 
& \textbf{43.82} & \textbf{18.91} 
& \textbf{45.25}	& \textbf{16.92}
& 80.61	& 75.37
\\

RaFe$_{(KTO)}$                  
& 62.12 & \textbf{28.12}
& \textbf{51.61} & \textbf{32.71} 
& 79.51 & \textbf{51.12} 
& \underline{43.27} & \underline{18.28} 
& \underline{45.03}	& \underline{16.40}
& 81.17	& 76.98
\\

\midrule
\multicolumn{13}{c}{\textsc{Expand}-Raw} \\
\midrule

LLM-Rewrite             
& 61.17 & 27.52
& 51.56 & 31.79 
& 80.20 & 50.29 
& 44.50 & 18.01 
& 45.13	& 16.98
& 81.30	& 78.12
\\

Query2Doc              
& 61.46 & 27.64 
& 50.75 & 30.83 
& 80.54 & 50.04 
& 44.49 & 18.75 
& 46.68	& 17.44
& 81.33	& \underline{79.48}
\\

SFT$_{(T_{\text{sft}})}$                   
& 62.01 & 26.76 
& 50.13 & 30.63 
& 80.42 & 50.21 
& 44.93 & 18.78 
& 47.15	& 17.82
& 81.26	& 71.95
\\

SFT$_{(T_{\text{all}})}$                 
& 62.21 & 26.36 
& 51.79 & 31.45 
& 80.57 & 50.24 
& 44.89 & 18.99 
& 47.51	& 17.54
& 81.49	& 72.48
\\

RaFe$_{(PPO)}$                  
& \underline{62.43} & \underline{28.31}
& 51.63 & 31.81 
& 80.32 & 50.01 
& 45.28 & 18.87 
& \underline{47.53}	& \textbf{18.22}
& \textbf{82.45}	& \textbf{80.15}
\\

RaFe$_{(DPO)}$                 
& 62.39 & 28.16
& \underline{52.30} & \underline{32.53}
& \underline{80.64} & \underline{50.92} 
& \underline{45.59} & \underline{19.25} 
& 47.25	& 17.92
& 81.73	& 78.85
\\

RaFe$_{(KTO)}$                  
& \textbf{62.65} & \textbf{28.50}
& \textbf{52.48} & \textbf{32.58}
& \textbf{80.88} & \textbf{51.24} 
& \textbf{45.91} & \textbf{19.52} 
& \textbf{47.93}	& \underline{18.11}
& \underline{82.16}	& 77.66
\\

\bottomrule
\end{tabular}
}
\caption{
The results showcase the performance in \textsc{Substitute}-Raw and \textsc{Expand}-Raw settings. ``QA'' refers to results obtained by Qwen-max, and ``w/o retrieval'' denotes generating answers directly. Results surpassing the OQR are highlighted in bold to represent the best-performing, while those underlined indicate the second-best.}
\label{tab:maintable_raw}
\end{table*}

%% file: sec/experiment.tex
\input{tab/main_table_ranked}

As we attempt to improve query rewriting for better RAG, we conduct our experiments on the typical RAG scenarios, Open-Domain Question Answering (ODQA). 
The process of RAG for ODQA can be formulated as $\mathcal{F}([D:q])$, where $\mathcal{F}$ denotes the LLMs, $q$ is the original query from datasets and $D$ is the documents concatenated for augmentation.

\subsection{Dataset}\label{sec:dataset}

To comprehensively validate the effectiveness and generalizability of our method, we conduct cross-lingual experiments. Specifically, we evaluate ReFe on both English and Chinese datasets.

\paragraph{English Datasets} For English data, we use several open-domain QA datasets including NQ~\citep{natural_question}, TriviaQA~\citep{triviaqa}, HotpotQA~\citep{hotpotqa}. For NQ and TriviaQA, we follow the split from previous work~\citep{dpr}, and default split for HotpotQA\footnote{\url{https://huggingface.co/datasets/hotpot_qa/viewer/fullwiki}}. 
We randomly gather 60k instances from the training set of the three datasets to conduct $T_{\text{all}}$ for training rewrite models.
As for evaluation, we collect the test set of NQ and TriviaQA, and the development set of HotpotQA as the held-in evaluation datasets.
Additionally, we use FreshQA~\citep{FreshQA}  for out-of-domain evaluation.

\paragraph{Chinese Datasets} 
For Chinese data, we gather a bunch of open-source queries to conduct the query set, the sources are listed in~\ref{tab:trainset_table}.
We use WebQA~\citep{WebQA_baidu} for the in-domain evaluation, while
FreshQA~\citep{FreshQA} (translated) for the out-of-domain evaluation. 
The process of translation can be found in Appendix~\ref{appendix:dataset}.

\subsection{Evaluation Settings}

In practical retrieval scenarios, query rewriting is commonly used as a technique to expand the retrieved documents based on the original query, followed by a re-ranking of the expanded documents. 
Thus, we validate \ours~in two experimental settings.
\paragraph{\textsc{Substitute}} Directly use the documents $D'$  retrieved by rewrite $q'$ for evaluation instead of the documents $D$ retrieved by query $q$.
\paragraph{\textsc{Expand}} Employing both $D$ and $D'$ for evaluation. We generate two rewrites $q_{0}', q_{1}'$ for the \textsc{Expand} setting with their retrieved $D'_{0}, D'_{1}$.

To further simulate the role of query rewriting in real-world scenarios, our experiments also include the performance under two following settings:

\paragraph{Raw} Concatenating top-5 retrieved documents in the default order. For \textsc{Expand} setting, the raw documents order is determined by sequentially and cyclically selecting the top documents from $D, D_{0}', D_{1}'$.
\paragraph{Ranked} Concatenating top-5 documents after re-ranking all the retrieved documents. As regard to \textsc{Expand} setting, all retrieved documents from both the query and rewrites are merged for ranking.

We utilize the Exact Match (EM) metric to evaluate the general QA performance.
Especially, we use Rouge-L~\citep{Rouge} to evaluate the \textit{false premise} set in FreshQA.
Given our work focus on open-domain QA, there are no gold documents or relevant annotations, 
we evaluate the retrieval by determining whether the retrieved documents contain the correct answer.
We report the Precision@K and the mean reciprocal rank (MRR) in the results.

\subsection{Baseline}

\paragraph{Original Query Retrieval (OQR)} Retrieve with the original query and utilize the documents by the default returned ranking from the search engine.

\paragraph{LLM Rewrite} Directly enable the LLMs to rewrite the original query with a few-shot prompt. In our experiment, we prompt Qwen-max to rewrite the original query.

\paragraph{Query2Doc} \citep{query2doc} A method creates pseudo-documents through few-shot prompting of LLMs and then the query is expanded with the generated pseudo-documents for retrieving. The used prompts are shown in Appendix~\ref{appendix:prompt}.

\paragraph{SFT} Use the pre-generated rewrites to directly train the rewrite model.  SFT${_{(T_{\text{sft}})}}$ represents the rewrite model trained specifically on the $T_{\text{sft}}$, while SFT${_{(T_{\text{all}})}}$ denotes the model trained on $T_{\text{all}}$.


\subsection{Implementation}

\paragraph{Retriever} We use an anonymous internal search engine for open domain to retrieve documents for the Chinese datasets, and Google Search for the English datasets. Specifically, we utilize the title and the summary snippet of the searched page as the retrieved documents for retrieval augmentation.

\paragraph{Base Model} We employ Qwen-max\footnote{\url{https://help.aliyun.com/zh/dashscope/developer-reference/api-details?spm=a2c4g.11186623.0.0.3d4a140b0kf3sd}}~\citep{qwen} to generate responses and 
conduct the evaluation with Qwen1.5-32b-chat. 
Query rewriting models are trained with the Qwen-7b-base.

\paragraph{Reranker} For a general RAG task like open-domain QA, 
we believe that if our approach yields positive results with a general reranker, when transferring to a specific domain (where a domain-specific reranker is available), it will perform even better.
Thus, we employ a publicly available bge-reranker\footnote{\url{https://huggingface.co/BAAI/bge-reranker-base}}~\citep{c-pack} to conduct open-domain QA experiments, which serves to demonstrate the effectiveness of the methods we designed.



%% file: tab/main_table_ranked.tex
\begin{table*}[ht]
\centering
\resizebox{2\columnwidth}{!}{
\begin{tabular}{l|cccccccc|cccc}

\toprule
\multirow{3}{*}[-0.5em]{\textbf{Method}} &  \multicolumn{8}{c|}{\textbf{EN}} & \multicolumn{4}{c}{\textbf{ZH}} \\
\cmidrule{2-13}
& \multicolumn{2}{c}{\textbf{FreshQA}} & \multicolumn{2}{c}{\textbf{NQ}} & \multicolumn{2}{c}{\textbf{TriviaQA}} & \multicolumn{2}{c|}{\textbf{HotpotQA}} & \multicolumn{2}{c}{\textbf{FreshQA}} & \multicolumn{2}{c}{\textbf{WebQA}} \\
& QA & Prec@5 & QA & Prec@5 & QA & Prec@5 & QA & Prec@5 & QA & Prec@5 & QA & Prec@5 \\

\midrule
OQR               
& 62.56 & 30.88
& 51.50 & 35.68 
& 80.17 & 52.57 
& 43.21 & 18.32 
& 44.67	& 17.27
& 81.37 & 78.27 
\\
\midrule
\multicolumn{13}{c}{\textsc{Substitute}-Ranked} \\
\midrule

LLM-Rewrite             
& 59.24 & 27.34
& 49.75 & 32.27 
& 78.53 & 50.43 
& 41.48 & 16.37 
& 42.85	& 16.26
& 80.53	& 76.32
\\

Query2Doc              
& 58.84 & 28.32
& 45.62 & 30.59 
& 77.26 & 52.01 
& 42.26 & 17.73 
& 43.81	& 16.61
& 81.22	& \textbf{79.92}
\\

SFT$_{(T_{\text{sft}})}$                   
& 60.69 & 28.42
& 50.99 & 34.01 
& 78.35 & 50.19 
& 42.26 & 17.64 
& 43.44	& 16.56
& 77.72	& 74.65
\\

SFT$_{(T_{\text{all}})}$                 
& 61.42 & 28.40
& 50.93 & 32.54 
& 78.15 & 50.33 
& 42.66 & 17.88 
& 44.40	& 16.20
& 78.16	& 75.61
\\

RaFe$_{(PPO)}$                  
& \textbf{63.01} & 30.56
& 51.26 & 34.61 
& 98.86 & 51.33 
& 42.57 & 18.45 
& 43.77	& 16.79
& \textbf{81.46}	& 76.90
\\

RaFe$_{(DPO)}$                 
& \underline{62.89} & 30.28
& \textbf{51.97} & \textbf{35.89} 
& \textbf{80.41} & \textbf{53.54} 
& \underline{43.77} & \underline{19.07}
& \textbf{45.49}	& \textbf{17.58}
& 80.53	& 76.37
\\

RaFe$_{(KTO)}$                  
& 62.71 & \textbf{31.00}
& \underline{51.86} & 35.62 
& \underline{80.23} & \underline{53.09} 
& \textbf{44.77} & \textbf{19.82} 
& \underline{45.30}	& \underline{17.36}
& 81.14	& 77.98
\\

\midrule
\multicolumn{13}{c}{\textsc{Expand}-Ranked} \\
\midrule

LLM-Rewrite             
& 62.34 & 31.14
& 51.55 & 36.34 
& 80.79	& 54.93
& 45.73	& 20.85
& 45.83	& 17.52
& 82.29	& 78.21
\\

Query2Doc              
& 63.06 & 31.84
& 51.83 & 37.16
& 80.28	& 54.47
& 45.82	& \underline{23.05}
& 46.58	& 18.29
& \underline{83.35}	& \underline{80.75}
\\

SFT$_{(T_{\text{sft}})}$                   
& 63.16 & 31.56
& 51.75 & 37.44 
& 80.17	& 54.20
& 45.18	& 22.28
& 47.61	& 18.86
& 82.08	& 79.15
\\

SFT$_{(T_{\text{all}})}$                 
& 63.27 & 28.44
& 51.94 & 37.68 
& 80.88	& 54.25
& 45.84	& 22.09
& 46.95	& 18.63
& 82.75	& 79.43
\\

RaFe$_{(PPO)}$                  
& \textbf{64.96} & \underline{33.54}
& 52.36 & \underline{38.44}
& 81.38	& 55.27
& \underline{46.73} & 22.39 
& \underline{48.83}	& \textbf{19.66}
& \textbf{83.58}	& \textbf{80.93}
\\

RaFe$_{(DPO)}$                 
& 63.98 & 33.20
& \underline{52.74} & \textbf{38.57} 
& \underline{81.74} & \underline{55.60}
& 46.53 & 22.78 
& 48.72	& 18.58
& 83.04	& 79.83
\\

RaFe$_{(KTO)}$                  
& \underline{64.85} & \textbf{33.72}
& \textbf{52.86} & 38.37 
& \textbf{81.97} & \textbf{55.67} 
& \textbf{46.79} & \textbf{23.35} 
& \textbf{48.96}	& \underline{19.25}
& 82.96	& 79.52
\\

\bottomrule
\end{tabular}
}
\caption{Results of \textsc{Substitute}-Ranked and \textsc{Expand}-Ranked settings. ``OQR'' is evaluated after ranking.}
\label{tab:maintable_ranked}
\end{table*}

%% file: sec/result.tex
\subsection{Main Result}




\input{tab/feedback_compare}

From Table~\ref{tab:maintable_raw} and Table~\ref{tab:maintable_ranked}, we can observe that RaFe outperforms other query rewriting baselines and OQR across almost all settings in retrieval and question-answering metrics.
It can be noted that the performances of most methods decrease slightly compared to OQR under the \textsc{Substitute} setting, where RaFe also shows marginal improvements.
The weak performance might be attributed to that rewriting tend to deviate from the original query in some challenging cases. We provide a deeper analysis in the Appendix~\ref{appendix:weak_performance}.

While under the \textsc{Expand} setting, the majority of baseline methods perform better than under \textsc{Substitute} setting. 
Notably, RaFe achieves significant improvements in the Expand-Ranked setting, where the QA results surpass all other baselines including OQR by 2\%-3\%. A similar conclusion can be drawn from Table~\ref{tab:qwen_result}. 
By comparing results between Table~\ref{tab:maintable_raw} and Table~\ref{tab:maintable_ranked}, it can be found that even with feedback provided to the query rewriting models through the use of rerankers, the ranked results continue to show a substantial increase in performance, which are further illustrated in Figure~\ref{fig:raw_ranked}.
It suggests that in practical applications of RAG, it may yield the greatest benefit by employing query rewriting with the \textsc{Expand}-Ranked setting. More retrieval results are shown in Appendix~\ref{appendix:retrieval_results}.

\subsection{Compared with Other Types of Feedback}

Previous work on training query rewrite models for the RAG~\citep{QueryRewrite_emnlp23} has leveraged LLMs performance on QA tasks as the feedback signal.
Many works construct feedback based on retrieval metrics from annotated documents~\citep{ConQRR, query_reformulation}. 
To thoroughly assess the efficacy of our approach, we also conduct experiment with these types of feedback. 
We obtain good-bad pairs (i.e. true for good and false for bad) for offline training introduced in Sec~\ref{sec:feedback training}. 
We use Qwen-32b-chat to conduct the LLM feedback. 
For the retrieval feedback, we utilize the results of Prec@5 to obtain good-bad pairs.
The results are shown in Table~\ref{tab:feedback_compare}. 
Additionally, we provide a comparison between reranker feedback and other feedback, demonstrated in Table~\ref{tab:feedback_cost}.

\input{tab/feedback_cost}

The results show that RaFe outperforms the other two types of feedback. Precision feedback yields the worst results, which may be attributed to the rudimentary construction of precision in our dataset—merely considering whether the answer is present within the document. 
LLM feedback also demonstrates competent performance in the \textsc{Substitute} setting. 
However, from Table~\ref{tab:feedback_cost}, we notice that under an equivalent data volume, the cost of employing LLM to construct feedback substantially exceeds that of the other two feedback.

%% file: tab/feedback_compare.tex
\begin{table}[ht]
\centering
\resizebox{0.95 \columnwidth}{!}{
\begin{tabular}{l|cccc}
\toprule
\multirow{2}{*}{\textbf{Methods}}  & \multicolumn{2}{c}{\textbf{FreshQA}} & \multicolumn{2}{c}{\textbf{NQ}}  \\
& Raw & Ranked & Raw & Ranked \\

\midrule
OQR       
& 61.87	& 62.56	& 51.36	& 51.50
\\

\midrule
\multicolumn{5}{c}{\textsc{Substitute}} \\
\midrule

SFT$_{(T_{\text{all}})}$  
& 60.55	& 61.42	& 50.39	& 50.93
\\

Precision$_{(DPO)}$
& 60.43	& 61.03	& 49.32	& 50.65
\\

Precision$_{(KTO)}$
& 60.54	& 61.34	& 49.76	& 50.12
\\

LLM$_{(DPO)}$ 
& 61.95	& \underline{62.45}	& 50.94	& 51.44
\\

LLM$_{(KTO)}$ 
& \textbf{62.32}	& 62.39	& \underline{51.34}	& \underline{51.54}
\\

RaFe$_{(KTO)}$  
& \underline{62.12}	& \textbf{62.71}	& \textbf{51.61}	& \textbf{51.86}		          
\\

\midrule
\multicolumn{5}{c}{\textsc{Expand}} \\
\midrule

SFT$_{(T_{\text{all}})}$  
& 61.42	& 63.27	& 51.79	& 51.94
\\
Precision$_{(DPO)}$
& 61.56	& 62.84	& 50.34	& 51.29
\\
Precision$_{(KTO)}$
& 61.79	& 63.15	& 50.69	& 51.32
\\
LLM$_{(DPO)}$ 
& \underline{62.43}	& 63.53	& 51.63	& 52.43
\\
LLM$_{(KTO)}$ 
& 61.87	& \underline{64.08} & \underline{51.89}	& \underline{52.23}
\\
RaFe$_{(KTO)}$  
& \textbf{62.65}	& \textbf{64.85}	& \textbf{52.48}	& \textbf{52.86}          
\\

\bottomrule
\end{tabular}
}
\caption{Results compared with different feedback, Precision and LLM indicates the retrieval feedback and LLM feedback, respectively.}
\label{tab:feedback_compare}
\end{table}

%% file: tab/feedback_cost.tex
\begin{table}[ht]
\centering
\resizebox{0.85 \columnwidth}{!}{
\begin{tabular}{l|ccc}
\toprule
\textbf{Methods} & \textbf{Feedback} & \textbf{Annotation} & \textbf{Cost} \\
\midrule
LLM         &   QA Results  & yes   & 78h \\
Precision   &   Retrieval   & yes   & 0.01h \\
RaFe        &   Reranker    & no    & 0.67h  \\
\bottomrule
\end{tabular}
}
\caption{The comparison of different types of Feedback. \textbf{Annotation} indicates the whether the labeled data is needed for the feedback signals. The \textbf{Cost} means the time for constructing the feedback for 30k instances.}
\label{tab:feedback_cost}
\end{table}

%% file: sec/analysis.tex
\begin{figure*}
    \centering
    \includegraphics[width=1\textwidth]{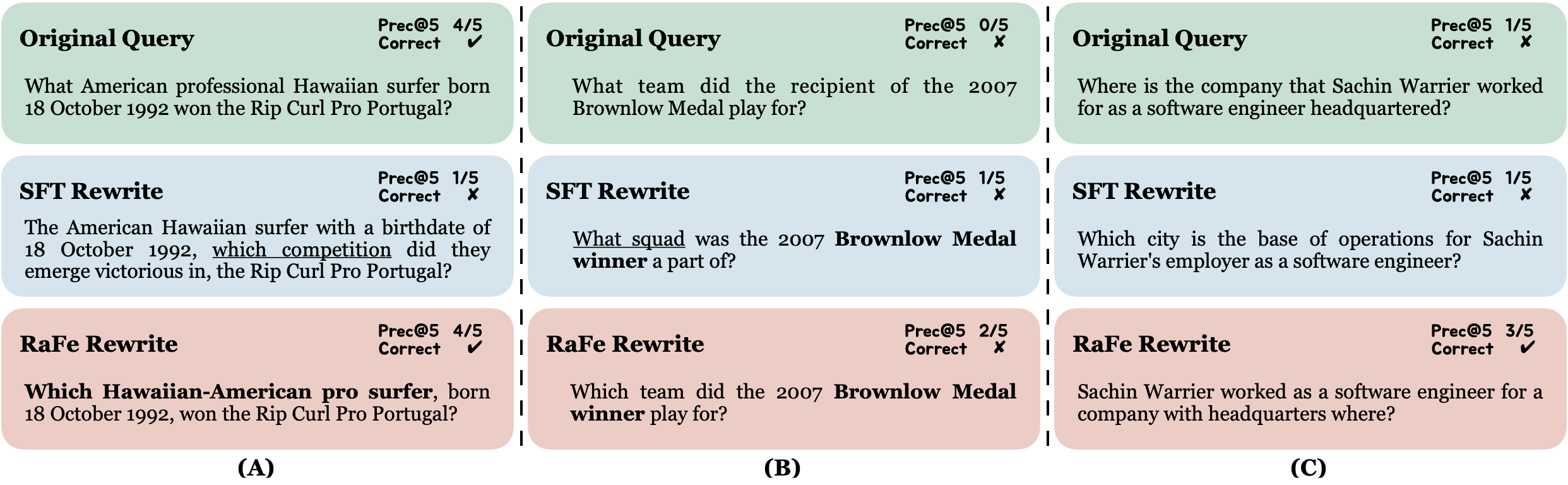}
    \caption{Three types of examples, including the original query and rewrites from SFT and RaFe.
    The Prec@5 results of queries and rewrites are presented, and \textbf{``Correct''} denotes that whether the prediction is correct or not.}
    \label{fig:case_full}
\end{figure*}

\subsection{How RaFe makes rewriting better?}

In this section, we present illustrative case studies to intuitively compare different rewrites and the original query in Figure~\ref{fig:case_full}.
The benifits of RaFe can be summarized into three types. 

\textbf{(A): RaFe performs better in preserving the semantics of the original query.} As shown in Figure~\ref{fig:case_full} (A), 
it can be observed that RaFe, after alignment through reranker, can rewrite queries in a way that better preserves the semantics of the original query. In contrast, the rewrite by SFT directly shifts the focus of the query from which athlete to which competition. 

\textbf{(B): RaFe's rewrites improve the format of the query for retrieval purposes.}
RaFe's rewrite is capable of transforming an uncommon term ``recipient'' into ``winner''. Although SFT rewrites also replace ``recipient'' with ``winner'', it changes ``team'' from a sports competition context to ``squad'', a term commonly used in military, police, or other contexts, thereby introducing potential ambiguity.

\textbf{(C): RaFe's rewrites sentences for better understanding.}
This kind of case is not easily discernible as good or bad based on intuition; however, RaFe's rewrite demonstrates better performance in retrieval results. Such cases show why we require feedback to enhance the QR effectiveness, as we always fail to articulate how a query could be formatted to better suit a retriever.

\begin{figure}
    \centering
    \includegraphics[width=0.48\textwidth]{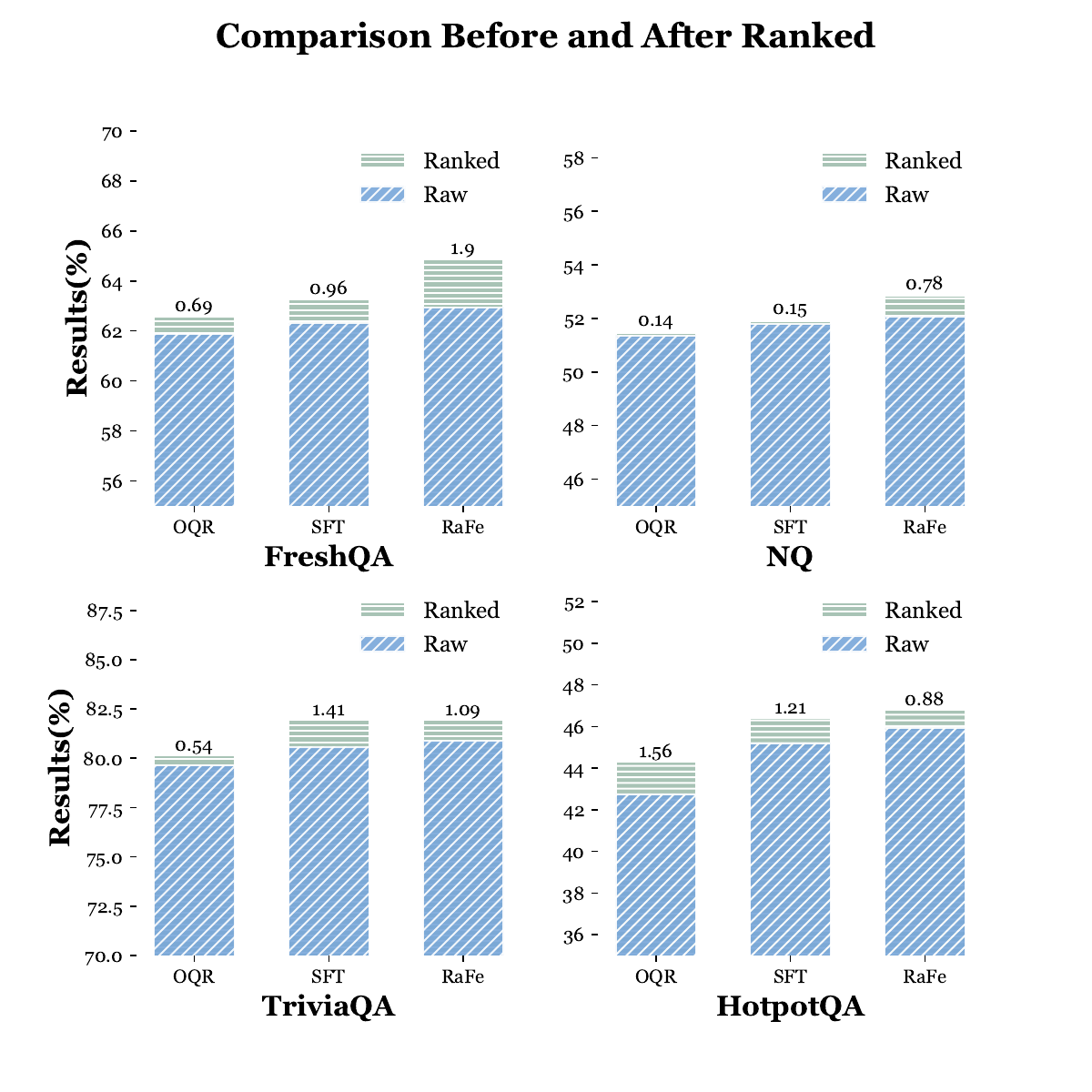}
    \caption{The performance of different rewrite models before and after all the documents are reranked under \textsc{Expand} setting. The number displayed on each bar represents the specific improvement from Raw to Ranked.}
    \label{fig:raw_ranked}
\end{figure}

\input{tab/prec_table}

\subsection{How does the Reranker Feedback Work?}



To investigate how reranker works for query rewriting, 
we first ascertain the ability of the publicly available reranker to rank on unseen datasets.

\begin{figure}
    \centering
    \includegraphics[width=0.5\textwidth]{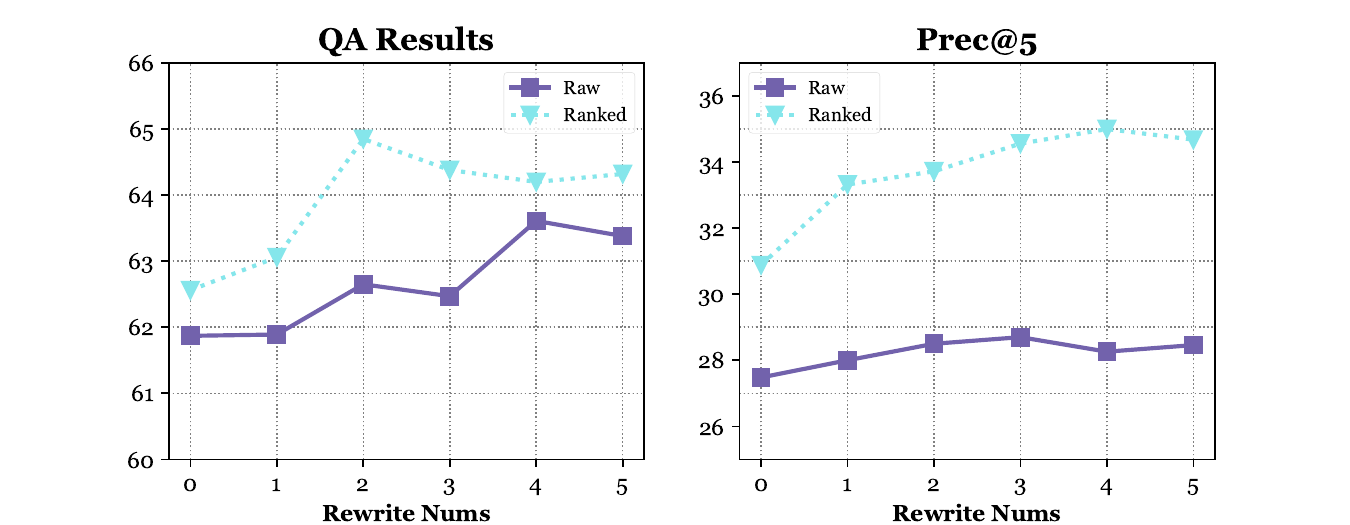}
    \caption{The results of different rewrite nums in \textsc{Expand} setting. We list the result from 0 to 5 rewrites. The rewrites are generate by RaFe$_{(KTO)}$.}
    \label{fig:rewrite_num}
\end{figure}
The comparing results are presented in Figure~\ref{fig:raw_ranked}. 
It can be clearly seen that all methods yield better QA performance after documents are ranked on all the datasets.
This indicates that the reranker's pattern for document sorting acts as a positive signal for the retrieval system. Meanwhile, we can observe that RaFe performs the better improvements after ranked, which further demonstrates the effectiveness of reranker feedback.




Moreover, we validate the effectiveness of reranker in constructing good and bad pairs within $T_{\text{f}}$. 
We compare the precision of documents retrieved by different queries in Table~\ref{tab:prec}. 
It is obvious that the documents retrieved by good rewrites exhibit significantly higher precision compared to those retrieved by the original query, which indicates that the reranker is capable of effectively distinguishing between rewrites that can retrieve high-quality documents and those that cannot. 
We also provide some examples in Appendix~\ref{appendix:good-bad cases}.

\subsection{How Many Rewrites is Optimal for RAG?}

In this section, we delve deeper into the impact that varying numbers of rewrites have on the final performance, since in practical applications of query rewriting, a balance must be struck between the quantity of generated rewrites and performance efficiency, given that generating more rewrites could potentially result in more response time.
We generate different numbers of rewrites, the results are depicted in Figure~\ref{fig:rewrite_num}.
The QA results peak when there are 4-5 rewrites, suggesting that employing more rewrites can yield considerable benefits by retrieving more relevant top documents. However, Prec@5 nearly approaches the best around 2-3 rewrites. 
When ranking all passages, the performance ceiling is attained with merely 2 rewrites. Considering the time cost, 2-3 rewrites may benefit the most for practical RAG.

%% file: tab/prec_table.tex
\begin{table}[ht]
\centering
\resizebox{0.85 \columnwidth}{!}{
\begin{tabular}{l|ccc}
\toprule
\textbf{Methods} & \textbf{Prec@5} & \textbf{Prec@10} & \textbf{MRR} \\
\midrule
Original Query   & 41.41          & 39.76   & 54.11 \\
Bad Rewrite      & 30.74          & 28.13   & 43.64 \\
Good Rewrite     & \textbf{46.14} & \textbf{44.34}  & \textbf{59.17} \\
\bottomrule
\end{tabular}
}
\caption{The comparison of retrieval results between original query and good/bad rewrites.}
\label{tab:prec}
\end{table}

%% file: sec/related_work.tex
\subsection{Query Rewriting}
Query rewriting is a critical technique within the retrieval domain~\citep{DBLP:journals/csur/CarpinetoR12, llm_for_ir_survey}.
With the groundbreaking advancements in scaling-up model capabilities, query rewriting has also played a pivotal role in enhancing the abilities of LLMs in RAG~\citep{dsp, self-ask, DBLP:journals/corr/abs-2401-15884}.
Many works~\citep{query2doc, lamer, informative_query_rewrite} directly leverage LLMs' strong capabilities to expand or rewrite queries.
Nonetheless, in practical application scenarios, a smaller rewriting model is preferred to avoid the costly requests for LLMs. At the same time, feedback training is the most commonly employed method to enhance the smaller rewriting models.
~\citet{query_reformulation} incorporates the ranking signals from annotated passages for better results, as well as previous works on conversational query rewrite~\citep{ConQRR, ConvGQR, reinforced_question_rewrite}. 
\citet{QueryRewrite_emnlp23} first generates answers from LLMs and then uses the QA evaluation results as the training signals.
\citet{taobao_rewrite} leverages search scoring functions intrinsic to the e-commerce framework to assess rewrite quality, informing feedback signals, which is exceedingly domain-specific, limiting its applicability to other domains.

These works depend on using particularly designed scores or annotated labels for feedback signals, while our proposed method can generically deliver feedback based on ranking results, without needing annotated passages.

\subsection{Learning From Feedback}
Recent advancements in Reinforcement Learning from Human Feedback (RLHF)~\citep{instruct_gpt} have been instrumental in aligning the generative capabilities of large models with human preferences, significantly prompting the creation of strong LLMs~\citep{gpt-4}.
Therefore, a large number of studies about feedback alignment have been emerging~\citep{secrets_1, secrets_2, dpo, rrhf, raft, kto}.
Some research efforts are concentrated on devising methods to provide new forms of feedback~\citep{rlaif, reflexion, self-refine, contemplation, crystal, rl4f, maf}.
\citet{reasons_to_reject} propose to train models from judgment language feedback.
\citet{tuna} designs two types of ranking feedback drawing from LLMs, to improve the performance.

Despite all these works, the exploration of feedback in rewriting is currently limited to direct feedback from LLMs~\citep{QueryRewrite_emnlp23} and domain-specific scoring~\citep{taobao_rewrite}. Such feedback approaches are costly and fail to utilize the effective signals from the IR system. 
While \citet{CodeRL} and \citet{RLTF} effectively leverage the feedback from Unit Test in the domain of code generation, we investigate more appropriate feedback signals for query rewriting in this paper, the reranker feedback.


%% file: sec/appendix.tex
\subsection{Feedback Training Loss}

\subsubsection{DPO Loss}
\begin{multline}
    \mathcal{L}_{dpo}
    = -\mathbb{E}_{(q, q^{'}_{g}, q^{'}_{b})\sim{T_f}} [ \log\sigma \\
    ( \beta\log\frac{\mathcal{M}_{\theta}(q^{'}_{g}|q)}{\mathcal{M}_{\text{ref}}(q^{'}_{g}|q)} - \beta\log \frac{\mathcal{M}_{\theta}(q^{'}_{b}|q)}{\mathcal{M}_{\text{ref}}(q^{'}_{b}|q)} ) ], 
\end{multline}\label{dpo_loss}
where $\beta$ is the temperature parameter for DPO, $\mathcal{M}_{\theta}$ is the rewrite model to be updated, and $\mathcal{M}_{\text{ref}}$ is the fixed model during the training phase.

\subsubsection{KTO Loss}\label{appendix:kto_loss}
The KTO~\citep{kto} (Kahneman-Tversky Optimization) method is based on \textit{prospect theory}~\citep{tversky1992advances}, which tells how human decides according to uncertain outcomes. 
The theory is proposed by the economists Kahneman \& Tversky. Compared to DPO, the training based on KTO only needs the signal that whether a rewrite $q^{'}$ is good or not, formulated as $(q,q^{'};\rho), \rho\in[\text{good, bad}]$. 
And the $\mathcal{L}_{kto}$ is computed as follows:

\begin{multline}\label{eq:kto_loss}
    \mathcal{L}_{kto}
    = \mathbb{E}_{(q,q^{'};\rho)\sim {T_f}}[w(q^{'})(1-\hat{h}(q,q^{'};\rho))] ,\\
    g(q,q^{'};\rho)= \beta\log \frac{\mathcal{M}_{\theta}(q^{'}|q)}{\mathcal{M}_{\text{ref}}(q^{'}|q)} - \\
    \mathbb{E}_{q^{'}\sim {T_f}}[\beta\textsc{KL}(\mathcal{M}_{\theta}||\mathcal{M}_{\text{ref}})] ,\\
    h(q,q^{'};\rho)= \left\{
    \begin{array}{ll}
        \sigma(g(q,q^{'};\rho)) & \text{if $\rho$ is good} \\
        \sigma(-g(q,q^{'};\rho)) & \text{if $\rho$ is bad}
    \end{array} \right. , \\
    w(q^{'})= \left\{
    \begin{array}{ll}
        \lambda_{good} & \text{if $\rho$ is good} \\
        \lambda_{bad} & \text{if $\rho$ is bad}
    \end{array} \right. .
\end{multline}

The default values for $\lambda_{good}$ and $\lambda_{bad}$ are set to 1. When there is an imbalance between the number of good and bad samples, specific values are determined as the following formula:
\begin{equation}
    \frac{\lambda_{\text{good}}n_{\text{good}}}{\lambda_{\text{bad}}n_{\text{bad}}} \in [1,\frac{4}{3}]
\end{equation}

\subsubsection{PPO Loss}\label{appendix:ppo_loss}
When implementing PPO training, we indicate the action $a_t$ at step $t$ as generating the next token $\hat{q}^{'}_t$, while the current state $s_t=(q, \hat{q}^{'}_{<t})$ is composed of the original query and generated rewrite tokens.
Here we directly use the ranking score as a reward, and by adding a KL-divergence regularization~\citep{DBLP:conf/iclr/RamamurthyABHSB23,DBLP:journals/csur/CarpinetoR12}, the reward is computed as follow:
\begin{equation}\label{eq:ppo_reward}
    R(s_t, a_t) = S_{\text{reranker}}(q^{'}|q)-\beta_{\text{KL}}\text{KL}(\mathcal{M}_{\theta}||\mathcal{M}_{\text{ref}})
\end{equation}
and then with a value network $V_{\phi}$ initialized from $\mathcal{M}_{\theta}$, the advantages function follows GAE~\citep{GAE} can be formulated as:
\begin{equation}
    \begin{split}
    & \delta_{t} = R(s_t, a_t)+V_{\phi}(s_{t}+1)-V_{\phi}(s_{t}), \\
    & A(s_t, a_t)=\sum\nolimits_{t^{'}=0}^{\infty}\lambda^{t^{'}}\delta_{t+t^{'}}
    \end{split}
\end{equation}
and the final objective function is composed of value loss and policy loss \cite{secrets_1}. 

\begin{equation}
    \begin{split}
        ~& \mathcal{L}_{\theta} = \mathbb{E}_{(s_t,a_t)\sim \mathcal{M}_{\theta}}[min( 
        \frac{\mathcal{M}_{\theta}(s_t,a_t)}{\mathcal{M}_{\text{ref}}(s_t,a_t)}A(s_t,a_t), \\
        ~& \text{clip}(\frac{\mathcal{M}_{\theta}(s_t,a_t)}{\mathcal{M}_{\text{ref}}(s_t,a_t)},
        1-\epsilon,1+\epsilon)A(s_t,a_t)
        )] , \\
        ~& \mathcal{L}_{\phi} = \mathbb{E}_{(s_t,a_t)\sim \mathcal{M}_{\theta}}(V_{\phi}(s_t)-R_{t})^{2}, \\
        ~& \mathcal{L}_{\text{ppo}} =  \mathcal{L}_{\theta} + \mathcal{L}_{\phi}
    \end{split}
\end{equation}

\subsection{Training Details}

\input{tab/trainset_table}

\subsubsection{Implementation} \label{appendix:training}
All model training are completed on a single machine with 4$\times$A100 GPUs. And the training prompt for the rewrite is listed in Table~\ref{tab:rewrite_prompt}.
\paragraph{SFT} We train the rewrite model with 2 epochs and set the learning rate to 5e-5.

\paragraph{PPO} The PPO implementation is carried out according to the TRL repo\footnote{\url{https://github.com/huggingface/trl}}\citep{vonwerra2022trl}.
In line with the empirical configurations in previous work~\citep{secrets_1}, we set the batch size to 32, and conduct the training for 1000 optimization steps, which is approximately equivalent to 1.067 epochs. The clip range parameter $\epsilon$, and the coefficient $\beta_{\text{KL}}$ for the KL divergence in Eq~\ref{eq:ppo_reward}, are both set to 0.2 as defaulted.

\paragraph{DPO \& KTO} We conduct the offline training for 1 epoch on all the good-bad rewrite data, with a learning rate of 5e-6. We set the temperature parameter $\beta$ to 0.1, following the default setting of the previous implementation\footnote{\url{https://github.com/ContextualAI/HALOs}}.

\subsubsection{Dataset Details}\label{appendix:dataset}

We list the sources and numbers of training instances in Table~\ref{tab:trainset_table}.

\paragraph{Initial Training Set of Rewrite Model}
For the open-domain QA task, we use qwen-max~\cite{qwen} to conduct the data production for both English and Chinese dataset.

\input{tab/retrieval_table_raw}

\paragraph{The Construction of Translated FreshQA}
We first translate the entire set of 500 FreshQA test questions,
and then manually review and filter each translation to identify those that were relatively more relevant to the Chinese internet. Ultimately, we obtained a set of 293 Chinese-translated FreshQA dataset.

\subsection{Additional Experimental Results}

\input{tab/qwen32b}

\subsubsection{The Retrieval Results}\label{appendix:retrieval_results}

We report complete retrieval results of Prec@10 and MRR in this section. The results of \textsc{Substitute}-Raw and \textsc{Expand}-Raw are shown in Table~\ref{tab:retrieval_raw}, while the results of \textsc{Substitute}-Ranked and \textsc{Expand}-Ranked are in Table~\ref{tab:retieval_ranked}.

\input{tab/retrieval_table_ranked}

Comparing the results between \textsc{Substitute} and \textsc{Expand}, it can be found that methods with lower retrieval results under the \textsc{Substitute} setting tended to show greater improvement under \textsc{Expand}. However, the retrieval results for RaFe do not exhibit great improvement under the \textsc{Expand}-Raw setting.
Further comparison between the QA results and retrieval metrics reveals that, generally, the trends of improvement in retrieval results align with those in QA performance. 

\subsubsection{QA Results of Qwen-32b}

\begin{figure}
    \centering
    \includegraphics[width=0.45\textwidth]{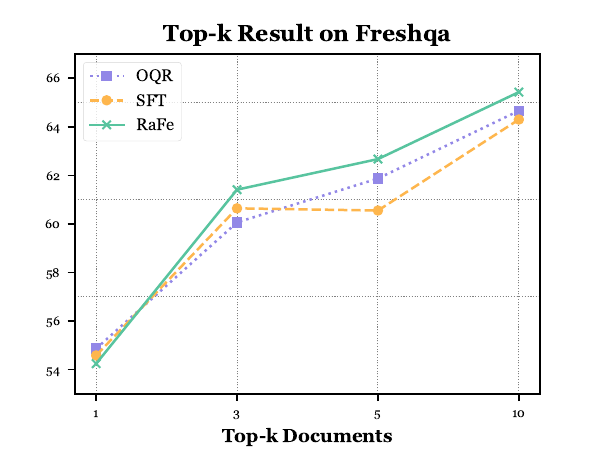}
    \caption{The results under \textsc{Substitute} setting on FreshQA with different number of documents.}
    \label{fig:topk_line}
\end{figure}

\input{tab/nq_prec_table}

To further demonstrate the results of our methods, we conduct experiments on different sizes of models. Specifically, we choose Qwen1.5-32b-chat for evaluation. The results are shown in Table~\ref{tab:qwen_result}.
The results indicate that RaFe consistently outperforms across almost all settings.
 Moreover, it is observed that compared to Qwen-max for QA, the 32B model exhibits lower performances.

It is worth noting that in the \textsc{Substitute}-Raw setting of the NQ dataset, utilizing Qwen-max does not yield great results. However, a significant improvement can be observed with Qwen-32b. 
This may suggest that for some cases beyond the capability coverage of qwen-32b, query rewriting can benefit the retrieval augmentation. 
As models increase in size, their inherent capabilities may become sufficient to handle these cases effectively, negating the need for query rewriting.

\subsubsection{Top-k Documents Results}
Additionally, we explore the performance of our proposed method when concatenating a different number of documents. 
We carry out the experiment on the Chinese version of the FreshQA.
The results presented in Figure~\ref{fig:topk_line} reveal that when solely the first document is utilized, the retrieval using the original query yields the best results.
As the number of concatenated documents increases, RaFe consistently outperforms both SFT and the original query results.

\subsection{Additional Analysis}

\subsubsection{The Relatively Weak Performance}\label{appendix:weak_performance}

From the results, it can be observed that there are only marginal improvements in some datasets, especially in \textsc{Substitute}-Raw setting. Taking the NQ dataset as an example, we attempt to investigate the difference between. The NQ dataset is a quite hard dataset, so for challenging cases, the minor reformulation of key phrases could cause the wrong retrieval. For instance, comparing Original Query: \textit{``what is the cross on a letter t called?''} and RaFe Rewrite: \textit{``What do you call the cross-like symbol on a letter 't'?''},
it can be found in the original query explicitly using ``cross on a letter t'' to a specific term related to typography. The rewritten query adds complexity and potential vagueness with a ``cross-like symbol'', which may mislead search engines towards broader symbol recognition or confuse with other types of crosses, thereby reducing the precision of the search results.

Additionally, the results on smaller models revealed that RaFe could achieve noteworthy improvements even in \textsc{Substitute}-Raw results. Thus, we obtain the cases answered both correctly and wrong by different size models.
As shown in Table~\ref{tab:nq_prec}, the average prec@5 on `good' cases is comparable between models of different sizes.
However, in `bad' cases, smaller models exhibit higher average precision. In contrast, when comparing the the results between Qwen-max and Qwen-32b, the improvements from RaFe diminish. This suggests that the benefits RaFe brings in simple cases are reduced as the model's parameter increases. Meanwhile, the deviations in more challenging cases are retained, which could lead to less impressive results. This further implies that query rewriting for RAG might be better suited for the \textsc{Expand} setting, to broaden the scope of the query to increase the chances of retrieving relevant information.

\subsubsection{Good-Bad Pairs Cases}\label{appendix:good-bad cases}

\begin{figure}[h]
    \centering
    \includegraphics[width=0.45\textwidth]{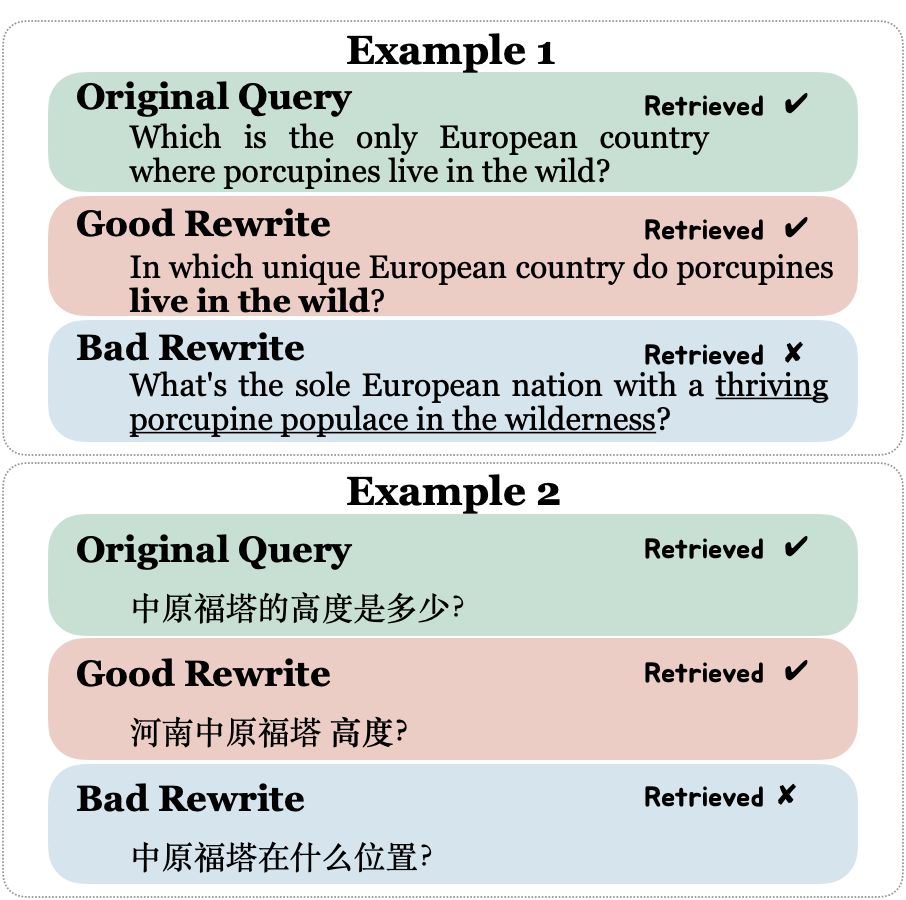}
    \caption{Two examples of good-bad rewrite pairs, each contains an original query, the good rewrite and bad rewrite. The \textbf{``Retrieved''} sign indicates whether the top 5 documents contains the answer or not.}
    \label{fig:case_2}
\end{figure}

In this section, we delve deeper into how rerankers take effect by presenting case studies. 
We investigate cases of how rerankers distinguish between good-bad pairs.
Figure~\ref{fig:case_2} provides two examples.

In the first example, the original query pertains to the only European country where wild porcupines reside. The good rewrite simplifies to a more direct question: ``In which unique European country do porcupines live in the wild?'' This rewrite is clear and precise. In contrast, the bad rewrite, ``What's the sole European nation with a thriving porcupine populace in the wilderness?'' Although conveying similar information, it appears excessively verbose and unnecessarily complex in its wording, resulting in failure in retrieval.

The second example's original query asks about the height of the Zhongyuan Pagoda in Henan Province, China. 
The good rewrite poses the same question in a more concise manner: ``Height of Zhongyuan Pagoda in Henan?''
This succinct rewrite may be better suited for rapid information retrieval. 
The bad rewrite, on the other hand, is: ``Where is the Zhongyuan Pagoda located?'' 
It fails to correctly rephrase the original question, as it shifts the focus from ``height'' to ``location'', causing a deviation from the original query's intent.
These cases demonstrate that the reranker's scoring of retrieved documents can effectively differentiate between good and bad rewrites.

\subsubsection{Additional Case for Better Format Rewriting}

\begin{figure}
    \centering
    \includegraphics[width=0.46\textwidth]{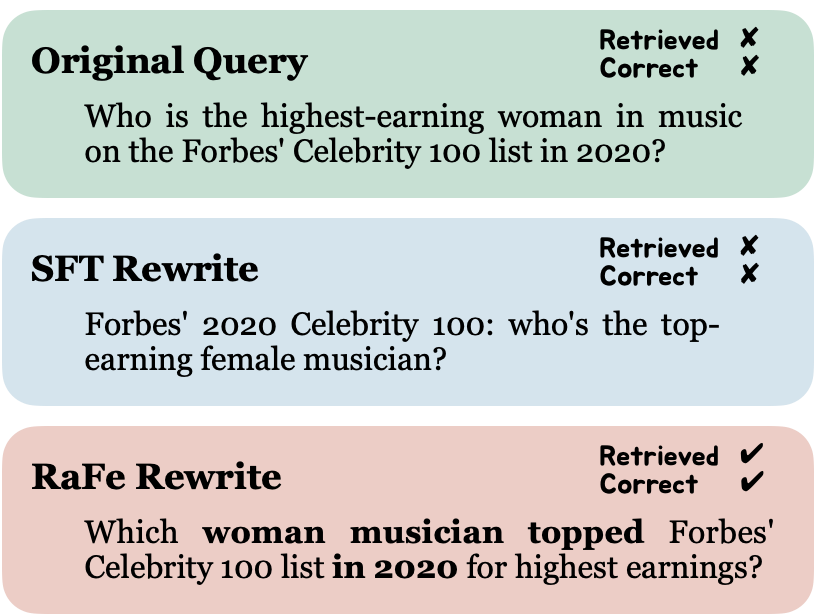}
    \caption{An example includes the original query and rewrite from SFT and RaFe. The The label \textbf{``Retrieved''} denotes whether the answer is present within the top 5 retrieved documents, and \textbf{``Correct''} denotes that whether the prediction is correct.}
    \label{fig:case_freshqa}
\end{figure}

We provide one more case in this section.
The original question used the phrase ``\textit{woman in music}'' to inquire about the highest-earning female musician in Forbes' Celebrity 100 list in 2020, which may not have been as intuitive for search engines, resulting in a failure to retrieve the documents. 
While the RaFe rewrite directly refines ``\textit{woman in music}'' into ``\textit{woman musician}'', rephrasing the question with vocabulary more suitable for retrieval purposes.

In contrast, the rewrite from SFT also conveys a clearer expression of ``\textit{female musician}'', 
but its format more closely resembles a headline or newspaper title, which may not be as suitable for a search query as a direct interrogative format. Additionally, it does not clearly express that the search is specifically for the year.






\subsection{Prompts}\label{appendix:prompt}
In this section, we list the prompt we used in this paper. The instruction prompt for rewrite model is shown in Table~\ref{tab:rewrite_prompt}, and the prompt for evaluation is in Table~\ref{tab:eval_prompt}.
The few-shot prompts used for Query2Doc are derived from ~\citet{query2doc}, and we use the same prompts for the LLMs Rewrite.

\input{tab/rewrite_prompt_table}

\input{tab/eval_prompt}

%% file: tab/trainset_table.tex
\begin{table}[ht]
\centering
\resizebox{0.55 \columnwidth}{!}{
\begin{tabular}{c|c|c}
\toprule
\textbf{Language} & \textbf{Source} & \textbf{Num} \\
\midrule
\multirow{6}{*}{\textit{ZH}}
&baike     &    6552    \\
&webqa     &    16486   \\ 
&sougouqa  &    9488    \\
&squadzen  &    6294    \\
&balle     &    9601    \\
&coig      &    15080   \\
\midrule
\multirow{3}{*}{\textit{EN}}
&hotpotqa     &  12471    \\
&triviaqa     &  28083    \\ 
&nq           &  19445    \\
\bottomrule
\end{tabular}
}
\caption{Data Source of the Training Instances for Open Domain QA.}
\label{tab:trainset_table}
\end{table}

%% file: tab/retrieval_table_raw.tex
\begin{table*}[ht]
\centering
\resizebox{2\columnwidth}{!}{
\begin{tabular}{l|cccccccc|cccc}

\toprule
\multirow{3}{*}[-0.5em]{\textbf{Method}} &  \multicolumn{8}{c|}{\textbf{EN}} & \multicolumn{4}{c}{\textbf{ZH}} \\
\cmidrule{2-13}
& \multicolumn{2}{c}{\textbf{FreshQA}} & \multicolumn{2}{c}{\textbf{NQ}} & \multicolumn{2}{c}{\textbf{TriviaQA}} & \multicolumn{2}{c|}{\textbf{HotpotQA}} & \multicolumn{2}{c}{\textbf{FreshQA}} & \multicolumn{2}{c}{\textbf{WebQA}} \\
& Prec@10 & MRR & Prec@10 & MRR & Prec@10 & MRR & Prec@10 & MRR & Prec@10 & MRR & Prec@10 & MRR \\

\midrule

OQR               
& 26.34 & 38.43
& 30.41 & 45.59
& 48.66 & 61.70
& 15.49 & 29.25
& 15.38 & 24.32 
& 74.97 & 84.48 
\\
\midrule
\multicolumn{13}{c}{\textbf{\textsc{Substitute}-Raw}} \\
\midrule

LLM-Rewrite             
& 24.31 & 35.18
& 27.27 & 41.74
& 46.35 & 59.89
& 13.44 & 25.86
& 15.64	& 22.00
& 73.86	& 83.87
\\

Query2Doc              
& 24.42 & 35.95
& 26.05 & 37.82
& \underline{48.71} & 59.24
& 14.96 & 25.54
& 14.98	& 24.23
& \textbf{75.77}	& \textbf{85.80}
\\

SFT$_{(T_{\text{sft}})}$                   
& 24.43 & 36.13
& 28.75 & 43.81
& 45.51 & 59.69
& 14.48 & 28.27
& 15.27	& 24.94
& 70.83	& 80.09
\\

SFT$_{(T_{\text{all}})}$                 
& 24.13 & 34.69
& 28.45 & 43.08
& 45.67 & 59.68
& 14.73 & 28.78
& 14.25	& 23.40
& 72.10	& 82.73
\\

RaFe$_{(PPO)}$                  
& 25.73 & 37.23
& 29.44 & 44.16
& 46.59 & 60.45
& 15.10 & 29.32
& \underline{15.44}	& \textbf{26.36}
& 72.47	& \underline{84.64}
\\

RaFe$_{(DPO)}$                 
& \underline{26.42} & \underline{28.75}
& 30.18 & 45.34
& 48.20 & \underline{61.91}
& \textbf{16.42} & \textbf{31.14}
& \textbf{16.20}	& 25.01
& 74.47	& 83.87
\\

RaFe$_{(KTO)}$                  
& \textbf{26.59} & \textbf{39.19}
& \textbf{30.78} & \textbf{45.92} 
& \textbf{48.86} & \textbf{62.09}
& 15.75 & \underline{29.93}
& \underline{15.65}	& \underline{25.97}
& 73.47	& 84.60
\\

\midrule
\multicolumn{13}{c}{\textbf{\textsc{Expand}-Raw}} \\
\midrule

LLM-Rewrite             
& 26.28 & 38.46
& 30.88 & 44.42
& 48.96 & 61.80 
& 16.25 & 28.72
& 16.24	& 24.79
& 76.27	& 86.09
\\

Query2Doc        
& 26.76 & 38.48 
& 29.99 & 44.77
& 48.78 & 60.44
& 17.15 & 30.18
& 17.51	& 25.80
& \textbf{77.93}	& 89.05
\\

SFT$_{(T_{\text{sft}})}$     
& 25.78 & 39.07
& 30.40 & 44.38
& 48.62 & 61.93
& 17.04 & 30.51
& 17.02	& \underline{26.64}
& 69.80	& 88.68
\\

SFT$_{(T_{\text{all}})}$                 
& 25.48 & 39.14
& 30.59 & 44.44
& 48.86 & 61.89
& 17.24 & \textbf{30.56}
& 16.62	& 25.75
& 70.35	& 88.86
\\

RaFe$_{(PPO)}$                  
& 27.12 & \underline{39.25}
& 30.46 & 45.42
& 48.67 & 61.73
& 17.24 & 30.41
& \textbf{17.82}	& 26.41
& \underline{76.21}	& \textbf{89.12}
\\

RaFe$_{(DPO)}$                 
& 26.98 & 38.85
& \underline{31.18} & 45.45
& \underline{49.63} & \underline{61.96}
& \underline{17.38} & 30.43
& 16.20	& 25.01
& 74.42	& 89.05
\\

RaFe$_{(KTO)}$                  
& \textbf{27.80} & \textbf{39.56}
& \textbf{31.22} & \textbf{45.73}
& \textbf{49.82} & \textbf{62.02}
& \textbf{17.67} & \underline{30.53}
& \underline{17.66}	& \textbf{26.86}
& 74.98	& \underline{89.10}
\\

\bottomrule
\end{tabular}
}
\caption{
The retrieval results of \textbf{\textsc{Substitute}-Raw} and \textbf{\textsc{Expand}-Raw} settings.}
\label{tab:retrieval_raw}
\end{table*}

%% file: tab/qwen32b.tex
\begin{table}[ht]
\centering
\resizebox{0.95\columnwidth}{!}{
\begin{tabular}{l|cccc}

\toprule
\multirow{2}{*}{\textbf{Method}} &  \multicolumn{2}{c}{\textbf{FreshQA}} & \multicolumn{2}{c}{\textbf{NQ}} \\
& Raw & Ranked & Raw & Ranked \\

\midrule
w/o retrieval
& 32.83 & -     
& 36.67 & - 
\\

OQR               
& 39.79 & 41.13
& 42.53 & 44.16 
\\
\midrule
\multicolumn{5}{c}{\textsc{Substitute}} \\
\midrule

LLM-Rewrite             
& 35.24 & 36.75
& 40.24	& 40.27
\\

Query2Doc              
& 34.97 & 35.63
& 40.05	& 41.32
\\

SFT$_{(T_{\text{sft}})}$                   
& 40.07 & 40.66
& 42.27	& 43.24
\\

SFT$_{(T_{\text{all}})}$                 
& 38.92 & 40.01
& 42.34	& 43.80
\\

RaFe$_{(PPO)}$                  
& \textbf{41.15} & \textbf{42.13}
& 42.57	& 44.23
\\

RaFe$_{(DPO)}$                 
& 38.18 & 39.73
& \underline{42.82} & \underline{44.84} 
\\

RaFe$_{(KTO)}$                  
& \underline{40.46} & \underline{41.77}
& \textbf{43.78} & \textbf{44.90 }
\\

\midrule
\multicolumn{5}{c}{\textsc{Expand}} \\
\midrule

LLM-Rewrite             
& 37.24 & 39.14
& 43.40	& 44.43
\\

Query2Doc              
& 38.78 & 39.29
& 44.13	& 45.07
\\

SFT$_{(T_{\text{sft}})}$                   
& 39.49 & 39.29
& 43.54	& 44.17
\\

SFT$_{(T_{\text{all}})}$                 
& 39.91 & 41.68
& 43.89	& 44.21
\\

RaFe$_{(PPO)}$                  
& 40.05 & \underline{42.64}
& 44.39	& 44.87
\\

RaFe$_{(DPO)}$                 
& \underline{40.41} & 42.37
& \underline{44.49}	& \underline{45.34}
\\

RaFe$_{(KTO)}$                  
& \textbf{40.74} & \textbf{43.79}
& \textbf{44.56}	& \textbf{45.64}
\\

\bottomrule
\end{tabular}
}
\caption{The QA results on Qwen1.5-32b-chat.}
\label{tab:qwen_result}
\end{table}

%% file: tab/retrieval_table_ranked.tex
\begin{table*}[ht]
\centering
\resizebox{2\columnwidth}{!}{
\begin{tabular}{l|cccccccc|cccc}

\toprule
\multirow{3}{*}[-0.5em]{\textbf{Method}} &  \multicolumn{8}{c|}{\textbf{EN}} & \multicolumn{4}{c}{\textbf{ZH}} \\
\cmidrule{2-13}
& \multicolumn{2}{c}{\textbf{FreshQA}} & \multicolumn{2}{c}{\textbf{NQ}} & \multicolumn{2}{c}{\textbf{TriviaQA}} & \multicolumn{2}{c|}{\textbf{HotpotQA}} & \multicolumn{2}{c}{\textbf{FreshQA}} & \multicolumn{2}{c}{\textbf{WebQA}} \\
& Prec@10 & MRR & Prec@10 & MRR & Prec@10 & MRR & Prec@10 & MRR & Prec@10 & MRR & Prec@10 & MRR \\

\midrule
OQR               
& 26.34 & 43.92
& 30.41 & 49.06
& 48.66 & 64.28
& 15.49 & 31.03
& 15.38 & 26.67
& 74.97 & 87.92 
\\
\midrule
\multicolumn{13}{c}{\textbf{\textsc{Substitute}-Ranked}} \\
\midrule

LLM-Rewrite             
& 24.31 & 40.31
& 27.27 & 47.28
& 46.35 & 62.79
& 13.44 & 27.20
& 15.64	& 24.53
& 73.86	& 85.23
\\

Query2Doc              
& 24.42 & 39.53
& 26.05 & 42.55
& \underline{48.71} & 61.18
& 14.96 & 27.24
& 14.98	& 25.29
& \textbf{75.77}	& \textbf{88.23}
\\

SFT$_{(T_{\text{sft}})}$                   
& 24.43 & 42.28
& 28.75 & 48.06
& 45.51 & 62.86
& 14.48 & 30.58
& 15.27	& \textbf{26.94}
& 70.83	& 80.71
\\

SFT$_{(T_{\text{all}})}$                 
& 24.13 & 41.17
& 28.45 & 47.87
& 45.67 & 62.94
& 14.73 & 30.19
& 14.25	& 21.39
& 72.10	& 80.26
\\

RaFe$_{(PPO)}$                  
& 25.73 & 43.14
& 29.44 & 48.52
& 46.59 & 63.52
& 15.10 & 30.60
& 15.44	& 26.15
& 72.47	& \underline{86.77}
\\

RaFe$_{(DPO)}$                 
& 24.42 & 43.19
& 30.18 & 48.97
& 48.20 & \underline{64.52}
& \textbf{16.42} & \underline{31.52}
& \textbf{16.20}	& 25.46
& 74.47	& 85.54
\\

RaFe$_{(KTO)}$                  
& \textbf{26.59} & 43.08
& \textbf{30.78} & \textbf{49.48}
& \textbf{48.86} & \textbf{65.17}
& \underline{15.75} & \textbf{32.28}
& \underline{15.65}	& \underline{26.50}
& 73.47	& 85.89
\\

\midrule
\multicolumn{13}{c}{\textbf{\textsc{Expand}-Ranked}} \\
\midrule

LLM-Rewrite             
& 29.45 & 42.14
& 32.42	& 48.97
& 52.14	& 64.78
& 18.32	& 32.06
& 18.02	& 26.32
& 77.23	& 87.12
\\

Query2Doc        
& 30.50 & 44.51
& 32.73	& 49.21
& 52.25 & 64.88
& 19.24	& 33.66
& 18.18	& 26.64
& \textbf{79.81}	& \textbf{88.81}
\\

SFT$_{(T_{\text{sft}})}$          
& 30.52 & 44.62
& 34.02	& 49.39
& 52.55	& 66.06
& 19.29	& 33.03
& \underline{18.87}	& 27.55
& 77.02	& 87.86
\\

SFT$_{(T_{\text{all}})}$         
& 23.71 & 41.31
& 34.36	& 49.64
& 52.65	& \underline{66.14}
& 19.34	& 33.22
& 18.16	& \underline{27.79}
& 77.21	& 87.90
\\

RaFe$_{(PPO)}$                
& 30.28 & \underline{44.29}
& 35.10	& \underline{50.37}
& 52.63	& 65.86
& 19.66	& \textbf{33.92}
& 18.56	& \textbf{29.17}
& \underline{79.26}	& \underline{88.47}
\\

RaFe$_{(DPO)}$     
& \underline{30.62} & 44.54
& \textbf{35.22}	& 50.10
& \textbf{53.55}	& 66.05
& \underline{19.77}	& \underline{33.81}
& 16.19	& 25.46
& 78.18	& 88.28
\\

RaFe$_{(KTO)}$                  
& \textbf{31.14} & \textbf{45.24}
& \underline{35.18} & \textbf{50.54}
& \underline{53.09}	& \textbf{66.46}
& \textbf{19.89}	& 33.75
& \textbf{18.90}	& 27.43
& 77.84	& 88.09
\\

\bottomrule
\end{tabular}
}
\caption{The retrieval results of \textbf{\textsc{Substitute}-Ranked} and \textbf{\textsc{Expand}-Ranked} settings.}
\label{tab:retieval_ranked}
\end{table*}

%% file: tab/nq_prec_table.tex
\begin{table}[ht]
\centering
\resizebox{0.8 \columnwidth}{!}{
\begin{tabular}{c|ccc}
\toprule
\textbf{Case Set} & \textbf{Model} & \textbf{OQR} & \textbf{RaFe} \\
\midrule
\multirow{2}{*}{Good} &
Qwen-max 
& 59.10 & 59.30 \\
&
Qwen-32b 
& 60.12 & 59.98 \\
\midrule
\multirow{2}{*}{Bad} &
Qwen-max 
& 5.37 & 5.73 \\
&
Qwen-32b 
& 11.21 & 11.69 \\
\bottomrule
\end{tabular}
}
\caption{The \textbf{Prec@5} results of NQ datasets answered by different size of Models under \textsc{Subsitute}-Raw setting. \textbf{Good} indicates the cases correctly answered by both OQR and RaFe, while \textbf{Bad} refers both uncorrect. 
}
\label{tab:nq_prec}
\end{table}

%% file: tab/rewrite_prompt_table.tex
\begin{table}[h]
\centering
\resizebox{1.0 \columnwidth}{!}{
\begin{tabular}{p{1.2\linewidth}}
\toprule
\textbf{Prompt}  \\
\midrule
Instruction: output the rewrite of input query \\
\\
Query: [ORIGINAL QUERY] \\
\\
Output: [TARGET]\\
\bottomrule
\end{tabular}
}
\caption{The instruction prompt for rewriting models, both training and inference.}
\label{tab:rewrite_prompt}
\end{table}

%% file: tab/eval_prompt.tex
\begin{table}[ht]
\centering
\resizebox{1.0 \columnwidth}{!}{
\begin{tabular}{p{1.2\linewidth}}
\toprule
\textbf{Prompt}  \\
\midrule

\textbf{USER}
\\
The following information may help answering questions: 
\\
<TOP-K DOCUMENTS>\\
\\
\textbf{LLMs}
\\
Sure, I have noted the information above. Is there anything I can assist you with or any questions I can help answer?\\
\\
\textbf{USER}\\
<QUESTION>\\

\bottomrule
\end{tabular}
}
\caption{The evaluation prompt when employing Qwen-max for open-domain QA.}
\label{tab:eval_prompt}
\end{table}